\documentclass[sn-mathphys-num]{sn-jnl}
\usepackage{graphicx}%
\usepackage{multirow}%
\usepackage{amsmath,amssymb,amsfonts}%
\usepackage{amsthm}%
\usepackage{mathrsfs}%
\usepackage[title]{appendix}%
\usepackage{xcolor}%
\usepackage{textcomp}%
\usepackage{manyfoot}%
\usepackage{booktabs}%
\usepackage{algorithm}%
\usepackage{algorithmicx}%
\usepackage{algpseudocode}%
\usepackage{listings}%
\usepackage{array}
\usepackage{booktabs} 
\usepackage{diagbox}
\usepackage{multirow}
\usepackage{booktabs}  
\usepackage{array} 
\usepackage{caption} 
\usepackage{setspace}
\usepackage{natbib}
\usepackage{geometry}
\usepackage[compact]{titlesec} 
\newcolumntype{P}[1]{>{\centering\arraybackslash}p{#1}}
\newcolumntype{M}[1]{>{\centering\arraybackslash}m{#1}}
\usepackage{multirow}  
\usepackage{subcaption}
\usepackage{tikz}
\usetikzlibrary{decorations.pathreplacing}
\theoremstyle{thmstyleone}%
\theoremstyle{thmstyletwo}%

\theoremstyle{thmstylethree}%

\begin{document}
	
	\title[Article Title]{Wave Network: An Ultra-Small Language Model}
	
	\author[1]{\fnm{Xin} \sur{Zhang}}\email{zha19053@ttu.edu}
	\author[2]{\fnm{Victor S.} \sur{Sheng}}\email{victor.sheng@ttu.edu}
	
	\affil[1]{\orgdiv{Department of Computer Science}, \orgname{Texas Tech University}, \orgaddress{\street{2500 Broadway}, \city{Lubbock}, \state{Texas}, \postcode{79409}, \country{USA}}}
	
	\affil[2]{\orgdiv{Department of Computer Science}, \orgname{Texas Tech University}, \orgaddress{\street{2500 Broadway}, \city{Lubbock}, \state{Texas}, \postcode{79409}, \country{USA}}}
		
	\abstract{We propose an innovative token representation and update method in a new ultra-small language model: the Wave network. Specifically, we use a \textbf{complex vector} to represent each token, encoding both global and local semantics of the input text. A \textbf{complex vector} consists of two components: a magnitude vector representing the \textit{global semantics} of the input text, and a phase vector capturing the \textit{relationships between individual tokens and global semantics}. Experiments on the AG News text classification task demonstrate that, when generating complex vectors from randomly initialized token embeddings, our single-layer Wave Network achieves 90.91\% accuracy with wave interference and 91.66\% with wave modulation---outperforming a single Transformer layer using BERT pre-trained embeddings by 19.23\% and 19.98\%, respectively, and approaching the accuracy of the pre-trained and fine-tuned BERT base model (94.64\%). Additionally, compared to BERT base, the Wave Network reduces video memory usage and training time by 77.34\% and 85.62\% during wave modulation. In summary, we used a 2.4-million-parameter small language model to achieve accuracy comparable to a 100-million-parameter BERT model in text classification.}
	\maketitle
	\section{Introduction}\label{sec1}
	
	Pre-trained token representations are crucial for many language processing models\cite{bib4}, both static token embedding methods, such as Skip-gram and Continuous Bag of Words (CBOW) \cite{bib1}, and context-dependent token embedding methods \cite{bib3}, along with representation updating mechanisms like attention \cite{bib5}, are core techniques in NLP.
	
	Despite the remarkable achievements, current NLP models face several inherent challenges. First, existing token embedding methods primarily focus on local semantics, lacking the ability to directly represent global semantics. Second, popular architectures like Transformer \cite{bib5} measures semantic similarity using the dot product between token vectors, which is computationally expensive. In multi-layer deep learning architectures, the dot product is computed in each attention head and layer, leading to significant resource demand in terms of computing power and time. Consequently, large language models (LLMs) require numerous layers and substantial hardware, data, and time resources to perform optimally on downstream tasks. 
	
	From a more fundamental perspective, \citet{bib6} suggests that natural language is a complicated signal system that conveys information with specific meanings through the production of sounds and the reception of hearing. The signal characteristics enable human language to express a wide range of abstract concepts. Based on this signal processing view, we focus on an ultra-small language model can approach or even surpass the performance of large language models on specific tasks.
	
	\section{Related Work}\label{sec2}
 	This research is closely related to multiple fields, including machine learning, natural language processing, and representation learning, which all rely on semantic processing. These fields include applications such as semantic search, knowledge graph, question answering, and various natural language processing techniques. Due to space limitations, we will focus on specific related research areas. For example, \citet{bib8} proposed a method to convert a token embedding space into a more understandable concept space, thereby reducing training costs and improving interpretability. \citet{bib7} proposed ELM (Embedding Language Model), which makes embeddings more interpretable and broadly applicable by injecting embeddings into LLMs, and then directly transform embedding vectors into understandable narratives. \citet{bib9} treats the hidden state in RNN (Recurrent Neural Network) \cite{bib18} as a dynamic, trainable machine learning model, and this model-based hidden state is updated during the training and testing phases through self-supervised learning rules. 
	 
	 In the Transformer architecture, representation update are carried out through the attention mechanism. Therefore, in addition to research on token embeddings, significant efforts have also been made to improve the attention mechanism itself. For instance, \citet{bib10} proposed a new and efficient sparse attention learning method based on the differentiable ranking of internal representations. \citet{bib11} proposed a block-level computation based on local attention to enhance the embedding method of context information. \citet{bib12} proposed the Perceiver, which leverages an asymmetric attention mechanism to iteratively distill inputs into a tight latent bottleneck, allowing it to scale to handle substantial inputs.
	
	\section{Method}\label{sec3}
	There are two main token embedding methods. The first one is fixed token embedding, represented by CBOW and Skip-gram, which cannot adapt to the dynamic meanings of tokens in varying contexts. The second is context-dependent embeddings, such as BERT, which generates different embeddings for the same token depending on its contexts. Because most current NLP methods, including Transformer, rely on dot products to measure local semantic similarity between token embeddings, they do not directly capture the global semantics of the entire input text.
	
	In contrast, our framework encodes both global and local semantics using complex vectors. Instead of relying on dot products for token representation updates, we apply addition or multiplication to complex vectors to simulate wave interference or modulation. This approach is possible because languages can be treated as complicated signal systems that convey meaning through sounds production and auditory reception \cite{fleisch2019student, towne2014wave}, and complex vectors correspond to physical wave representations \cite{berman2018introductory, herman2015introduction}.
	
	\subsection{Represent Tokens Using Complex Vectors}
	We use a \textbf{complex vector} $\mathbf{G} \cdot e^{i \cdot \boldsymbol{\alpha}}$ to represent each token, encoding both global and local semantics of the input text. A \textbf{complex vector} consists of two components: a magnitude vector $\mathbf{G}$ representing the \textit{global semantics} of the input text, and a phase vector $\boldsymbol{\alpha}$ capturing the \textit{relationships between individual tokens and global semantics}. This representation is referred to as \textbf{complex vector token representation}. The detailed computation of each component will be explained in the following subsections.
	
	\subsubsection{Global Semantics Representation}\label{sec:global_semantics}
	The meaning of each token in the input text often depends on the overall semantics of the input text. Understanding global semantics helps resolve the ambiguity of individual tokens, which is essential for many downstream tasks requiring a holistic view of the input text.
	
	From a signal processing perspective, \textit{signals are often represented in polar coordinates and processed in the frequency domain.} A signal's amplitude represents its strength, and its phase describes its position relative to another signal within a cycle \cite{Feynman_1494701}. Referring to signal processing, we treat token embeddings as discrete signals in the frequency domain. Given an input text with $n$ tokens $input\_ text = [\mathbf{w}_1, \mathbf{w}_2, \dots,\mathbf{w}_j, \dots,\mathbf{w}_{n}]$:
	\setlength{\abovedisplayskip}{3pt}  
	\setlength{\belowdisplayskip}{3pt}  
	\begin{figure}[htbp]
		\vspace{-20pt}  
		\centering
		\begin{equation}
			\begin{aligned}
				&	\textit{dimension 1}\;\;\;\; \textit{dimension k}\\[-5pt]
				&  \;\;\;\;\;\;\;\downarrow \;\;\;\;\;\;\;\;\;\;\;\;\;\;\;\;\;\;\;\;\;\;\downarrow \\[-5pt]
				\mathbf{w}_1 &= \left[w_{1,1}, w_{1,2}, \dots, w_{1,k},\dots, w_{1,768} \right] \\
				\mathbf{w}_2 &= \left[w_{2,1}, w_{2,2}, \dots, w_{2,k}, \dots, w_{2,768} \right] \\[-5pt]
				& \makebox[45pt][c]{\vdots}  \makebox[10pt][c]{\vdots}\makebox[60pt][c]{\vdots}\makebox[25pt][c]{\vdots}\\[-5pt]
				\mathbf{w}_j &= \left[w_{j,1}, w_{j,2}, \dots, w_{j,k}, \dots, w_{j,768} \right] \\[-5pt]
				& \makebox[45pt][c]{\vdots}  \makebox[10pt][c]{\vdots}\makebox[60pt][c]{\vdots}\makebox[25pt][c]{\vdots}\\[-5pt]
				\mathbf{w}_n &= \left[w_{n,1}, w_{n,2}, \dots, w_{n,k}, \dots, w_{n,768} \right]\\
				& \; \;\;\; \textbf{Token  Embedding  Matrix}
			\end{aligned}
			\hspace{1cm}
			\begin{aligned}
				&  \textit{dimension 1}\;\; \textit{dimension k}\\[-5pt]
				&  \;\;\;\;\downarrow \;\;\;\;\;\;\;\;\;\;\;\;\;\;\;\;\;\;\;\;\;\downarrow \\[-5pt]
				& \left[ G_{1,1}, G_{1,2}, \dots, G_{1,k},\dots, G_{1,768} \right] \\
				& \left[ G_{2,1}, G_{2,2}, \dots, G_{2,k}, \dots, G_{2,768} \right] \\[-5pt]
				& \makebox[30pt][c]{\vdots}  \makebox[20pt][c]{\vdots}\makebox[60pt][c]{\vdots}\makebox[25pt][c]{\vdots}\\[-5pt]
				& \left[ G_{j,1}, G_{j,2}, \dots, G_{j,k}, \dots, G_{j,768} \right] \\[-5pt]
				& \makebox[30pt][c]{\vdots}  \makebox[20pt][c]{\vdots}\makebox[60pt][c]{\vdots}\makebox[25pt][c]{\vdots}\\[-5pt]
				& \left[ G_{n,1}, G_{n,2}, \dots, G_{n,k}, \dots, G_{n,768} \right]\\
				& \; \;\; \;\;\;\textbf{Token Magnitude  Matrix}
			\end{aligned}
			\label{embedding}
			\vspace{-9pt}  
		\end{equation}
		\caption{Token Embedding Matrix and Magnitude Matrix}
		\begin{flushleft}
			\vspace{-5pt}  
			\footnotesize
			\textbf{Note:} \textit{The number of dimensions can vary. In this proposal, we follow BERT base and use 768 dimensions to facilitate understanding.}
		\end{flushleft}
		\label{fig:embedding_matrix}
		\vspace{-11pt}  
	\end{figure}
	
	Each token embedding \( \mathbf w_{j} \) can be treated as a discrete real-value signal, where each elements $w_{j,k}$ represents the signal component along the $k$-th dimension. From a physical perspective, the magnitude of each signal component is defined as $G_{j,k}=|{w}_{j,k}|$, and the energy of each signal component can be defined as $E = {w}_{j,k}^2$ \cite{10.5555/248702}. Using these magnitudes, we calculate the token magnitude matrix (i.e., the right sub-figure in Figure \ref{fig:embedding_matrix}) from the token embedding matrix (i.e., the left sub-figure of Figure \ref{fig:embedding_matrix}). 
	\begin{figure}[htbp]
		\vspace{-14pt}  
		\centering
		\begin{tikzpicture}
			\node at (0, 0) {
				\begin{minipage}{\textwidth}
					\begin{equation*}
						\begin{aligned}
							& \;\;\;\;\;\;\;\;\;\;\;\;\;\;\;\;\;\;\;\;\;\;\;\;\;\;\;\;\;\;\;\;\;\;\;\;\;\;\;\;\;\;\;\;\;\;\;\;\mathbf{G}\\[-15pt]
							& \;\;\;\;\;\;\;\;\;\;\;\;\;\;\;\;\;\;\;\;\;\;\rotatebox{270}{
								\begin{tikzpicture}
									\draw [decorate, decoration={brace,amplitude=10pt}, thin] (0,0) -- (0,5.5);
								\end{tikzpicture}
							}\\[-5pt]
							& \;\;\;\;\;\;\;\;\;\;\;\;\;\;\;\;\;\;\;\text{G}_1,\;\;\;\;\;\;\;\;\;\;\;\text{G}_2,\;\;\;\dots\;, \;\;\;\;\;\text{G}_k,\;\;\;\dots\;, \;\;\;\;\;\text{G}_{768} \\[-5pt]
							& \begin{array}{cccccccc}
								{
									\begin{array}{c}
										\mathbf{w}_1 = \\ 
										\; \\
										\mathbf{w}_2 = \\
										\; \\[-5pt]
										\vdots \\[-5pt]
										\; \\
										\mathbf{w}_j = \\
										\; \\[-5pt]
										\vdots \\[-5pt]
										\; \\
										\mathbf{w}_n =
									\end{array}
								}
								&
								\sqrt{
									\begin{array}{c}
										w_{1,1}^2 \\
										\; +\\
										w_{2,1}^2 \\
										\; +\\[-5pt]
										\vdots \\[-5pt]
										\; +\\
										w_{j,1}^2 \\
										\; +\\[-5pt]
										\vdots \\[-5pt]
										\; +\\
										w_{n,1}^2
									\end{array}
								}
								&
								\begin{array}{c}
									\;\\
									\;\\
									\;\\
									\;\\
									\;\\
									\;\\
									\;\\
									\;\\
									\;\\
									\;\\
									\;\\
									\;
								\end{array}
								&
								\sqrt{
									\begin{array}{c}
										w_{1,2}^2 \\
										\; +\\
										w_{2,2}^2 \\
										\; +\\[-5pt]
										\vdots \\[-5pt]
										\; +\\
										w_{j,2}^2 \\
										\; +\\[-5pt]
										\vdots \\[-5pt]
										\; +\\
										w_{n,2}^2
									\end{array}
								}
								&
								\begin{array}{c}
									\dots \\
									\; \\
									\dots \\
									\; \\[-5pt]
									\vdots \\[-5pt]
									\; \\
									\dots \\
									\; \\[-5pt]
									\vdots \\[-5pt]
									\; \\
									\dots
								\end{array}
								&
								\sqrt{
									\begin{array}{c}
										w_{1,k}^2 \\
										\; +\\
										w_{2,k}^2 \\
										\; +\\[-5pt]
										\vdots \\[-5pt]
										\; +\\
										w_{j,k}^2 \\
										\; +\\[-5pt]
										\vdots \\[-5pt]
										\; +\\
										w_{n,k}^2
									\end{array}
								}
								&
								\begin{array}{c}
									\dots \\
									\; \\
									\dots \\
									\; \\[-5pt]
									\vdots \\[-5pt]
									\; \\
									\dots \\
									\; \\[-5pt]
									\vdots \\[-5pt]
									\; \\
									\dots
								\end{array}
								&
								\sqrt{
									\begin{array}{c}
										w_{1,768}^2 \\
										\; +\\
										w_{2,768}^2 \\
										\; +\\[-5pt]
										\vdots \\[-5pt]
										\; +\\
										w_{j,768}^2 \\
										\; +\\[-5pt]
										\vdots \\[-5pt]
										\; +\\
										w_{n,768}^2
									\end{array}
								}
							\end{array}\\
						\end{aligned}
					\end{equation*}
				\end{minipage}
			};
		\end{tikzpicture}
		\caption{Global Semantic Vector}
		\label{fig:global_semantic_vector}
		\vspace{-18pt}  
	\end{figure}
	
	Next, we sum the magnitudes of all token embedding components along each dimension to define the \textbf{global semantics vector} $\mathbf{G} = [{G}_1, {G}_2, \dots, {G}_k, \dots, {G}_{768}]$, where each \textbf{global semantic element} ${G_k}$ can be defined as $\text{G}_{k} = \left\| \mathbf{w}_{:,k} \right\|_2=\left\| [w_{1,k}, w_{2,k}, \dots,w_{j,k} , \dots,w_{n,k}] \right\|_2= \sqrt{{w_{1,k}^2 + w_{2,k}^2 + \dots +w_{j,k}^2  + \dots+ w_{n,k}^2}}$, as shown in Figure \ref{fig:global_semantic_vector}. Here, \( w_{j,k} \) represents the \( k \)-th dimension of the \( j \)-th token embedding. This \textbf{global semantic vector} \( \mathbf{G} \) represents the global semantics of the entire input text and will serve as the magnitude of the \textbf{complex vector token representation} of each token in polar coordinates.
	
	For simplicity, we focus on input-level global semantics in this research. As shown in Figure \ref{fig:embedding_matrix}, given a input text with $n$ token embeddings, the input-level global semantics vector can be defined as $\mathbf{input\_G} = \left[{input\_G}_1,{input\_G}_2, \dots,{input\_G}_k, \dots, {input\_G}_{768} \right]$, where ${input\_G}_{k} = \sqrt{{w_{1,k}^2 + w_{2,k}^2 + \dots +w_{j,k}^2 +\dots+ w_{n,k}^2}}$.
	
	\subsubsection{Local Semantics Representation}\label{sec:local_semantics}
	Local semantics usually represent the specific meaning of individual tokens, which helps in analyzing dependencies and subtle differences between tokens in an input text. For example, tasks like sentiment analysis, entity recognition, and keyword extraction, often require a precise understanding of individual tokens.
	
	\noindent\textbf{1)} \textbf {Using phase to represent relationships between tokens and the global semantic vector}
	
	From a signal processing perspective, \textit{phase describes the relative relationships between signals. We will use the phase of \textbf{complex vector token representations} to represent the relative relationships between individual tokens and the \textbf{global semantic vector}.} That is, the phase representation of a token is coupled with its \textbf{global semantic vector}. For each token $\boldsymbol{w_{j}}$ in the input text, its \textbf{phase vector} is $\boldsymbol{\alpha_j} = \left[{\alpha_1}, {\alpha_2}, \dots,{\alpha_k}, \dots, {\alpha_{768}} \right]=\left[{input\_\alpha_1}, {input\_\alpha_2}, \dots,{input\_\alpha_k}, \dots, {input\_\alpha_{768}} \right]$, where $\mathit{input\_\alpha_k}$ is defined as $arctan2(\frac{\sqrt{1-(\frac{w_{j,k}}{{input\_G}_{k}})^2}}{\frac{w_{j,k}}{{input\_G}_{k}}})$ based on the corresponding element $\mathit{input\_G_k}$ in the \textbf{global semantic vector} of the input text $\mathbf{input\_G} = \left[{input\_G}_1,{input\_G}_2, \dots,{input\_G}_k, \dots, {input\_G}_{768} \right]$. Note that we use the function arctan2 to ensure angles fall within the range of $-\pi$ to $\pi$, consistent with the standard phase angle in physics. With these definitions, we can derive the token phase matrix (i.e., the right sub-figure in Figure \ref{fig:embedding_matrix2}) from the token embedding matrix (the left sub-figure of Figure \ref{fig:embedding_matrix2}). 
	\begin{figure}[htbp]
		\vspace{-20pt}  
		\centering
		\begin{equation}
			\begin{aligned}
				\mathbf{w}_1 &= \left[ w_{1,1}, w_{1,2}, \dots, w_{1,k},\dots, w_{1,768} \right] \\
				\mathbf{w}_2 &= \left[ w_{2,1}, w_{2,2}, \dots, w_{2,k}, \dots, w_{2,768} \right] \\[-5pt]
				& \makebox[45pt][c]{\vdots}  \makebox[10pt][c]{\vdots}\makebox[60pt][c]{\vdots}\makebox[25pt][c]{\vdots} \\[-5pt]
				\mathbf{w}_j &= \left[ w_{j,1}, w_{j,2}, \dots, w_{j,k}, \dots, w_{j,768} \right] \\[-5pt]
				& \makebox[45pt][c]{\vdots}  \makebox[10pt][c]{\vdots}\makebox[60pt][c]{\vdots}\makebox[25pt][c]{\vdots} \\[-5pt]
				\mathbf{w}_n &= \left[ w_{n,1}, w_{n,2}, \dots, w_{n,k}, \dots, w_{n,768} \right] \\
				& \; \;\;\; \textbf{Token Embedding Matrix}
			\end{aligned}
			\hspace{1cm}
			\begin{aligned}
				\boldsymbol{\alpha}_1 &= \left[ \alpha_{1,1}, \alpha_{1,2}, \dots, \alpha_{1,k},\dots, \alpha_{1,768} \right] \\
				\boldsymbol{\alpha}_2 &= \left[ \alpha_{2,1}, \alpha_{2,2}, \dots, \alpha_{2,k}, \dots, \alpha_{2,768} \right] \\[-5pt]
				& \makebox[45pt][c]{\vdots}  \makebox[10pt][c]{\vdots}\makebox[60pt][c]{\vdots}\makebox[25pt][c]{\vdots} \\[-5pt]
				\boldsymbol{\alpha}_j &= \left[ \alpha_{j,1}, \alpha_{j,2}, \dots, \alpha_{j,k}, \dots, \alpha_{j,768} \right] \\[-5pt]
				& \makebox[45pt][c]{\vdots}  \makebox[10pt][c]{\vdots}\makebox[60pt][c]{\vdots}\makebox[25pt][c]{\vdots} \\[-5pt]
				\boldsymbol{\alpha}_n &= \left[ \alpha_{n,1}, \alpha_{n,2}, \dots, \alpha_{n,k}, \dots, \alpha_{n,768} \right] \\
				& \; \;\;\;\;\;\;\; \textbf{Token Phase Matrix}
			\end{aligned}
			\vspace{-11pt}  
		\end{equation}
		\caption{Token Embedding Matrix and Phase Matrix}
		\label{fig:embedding_matrix2}
		\vspace{-20pt}  
	\end{figure}

	\noindent\textbf{2)} \textbf {Illustrate Complex Vector Token Representation in Cartesian coordinate system} \label{sec:cartesion}
	
\begin{figure}[H]
		\vspace{-20pt}
		\centering
		\includegraphics[width=0.9\textwidth]{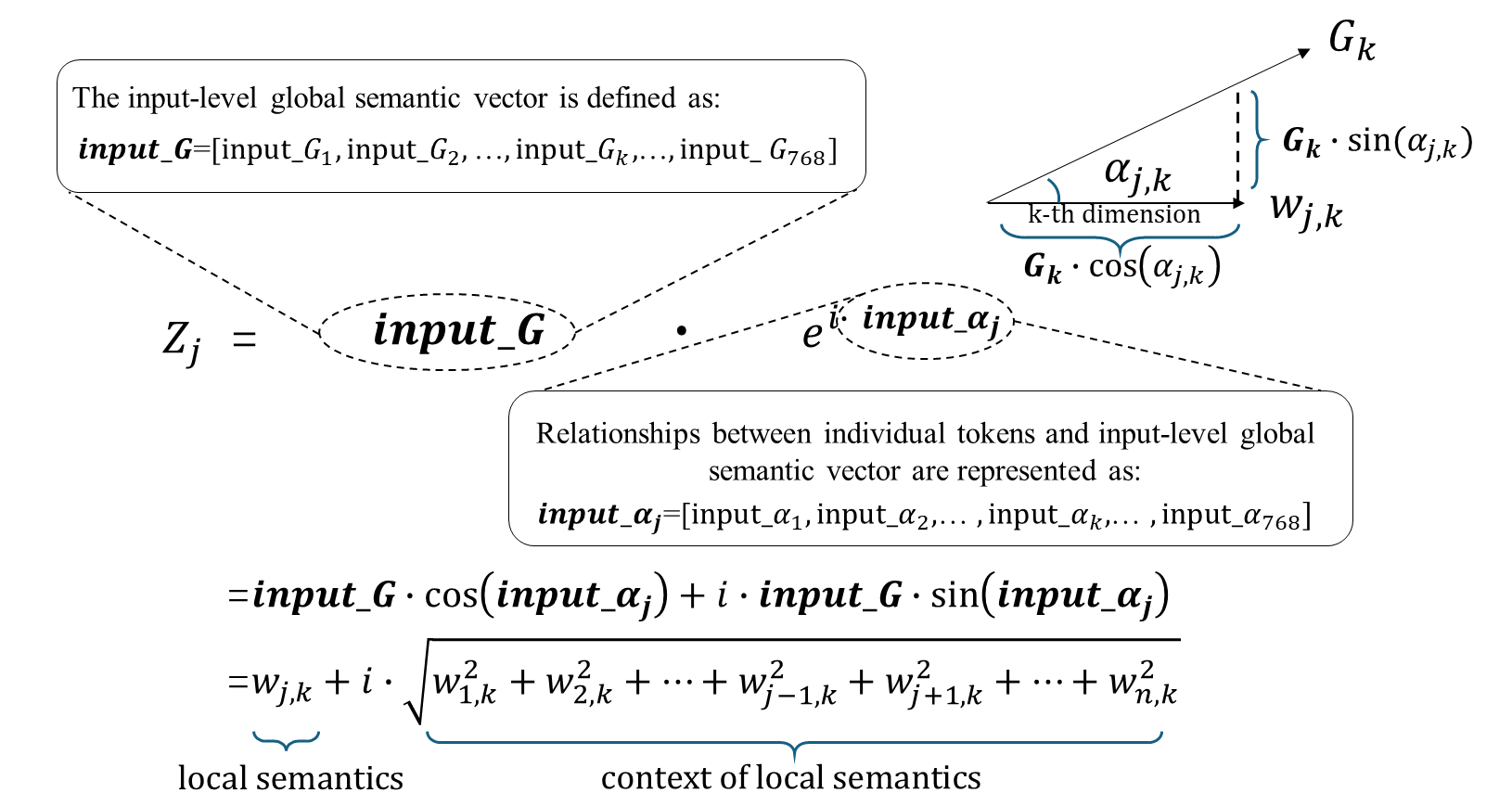}
		\caption{Convert \textbf{complex vector token representations} from polar coordinates to Cartesian coordinates}
		\label{angle}
		\vspace{-18pt}
	\end{figure}

	To illustrate the meaning of \textbf{complex vector token representations}' components in Cartesian coordinates, we use the Euler's formula $e^{i\theta} = \cos(\theta) + i \cdot \sin(\theta)$ \cite{Ahlfors1966} to convert \textbf{complex vector token representations} from polar to Cartesian coordinates, as shown in Figure \ref{angle}. For example, the \textbf{complex vector token representations} $\mathbf{input\_G}$ can be expressed in Cartesian coordinates as $\mathbf{input\_G} \cdot \cos(\boldsymbol{input\_\alpha_j}) + i \cdot \mathbf{input\_G} \cdot\sin(\boldsymbol{input\_\alpha_j})$. The inner product of $\sin(\boldsymbol{input\_\alpha_j})$ and $\cos(\boldsymbol{input\_\alpha_j})$ is zero over a full period, making them orthogonal \cite{arfken2013mathematical}. Consequently, the real part $\mathbf{input\_G} \cdot \cos(\boldsymbol{input\_\alpha_j})$ and the imaginary part $i \cdot \mathbf{input\_G} \cdot \sin(\boldsymbol{input\_\alpha_j})$ are also orthogonal, fulfilling the properties of wave representations as described in physics \cite{jackson_classical_1999}. As Figure \ref{angle} illustrates, the real part of the token embedding $\boldsymbol{w_{j}}$ represents the token's contribution along the \( k \)-th dimension, capturing the local semantics of the input text. The imaginary part describes the \textbf{global semantic element} apart from $\boldsymbol{w_{j}}$, representing the context of token $\boldsymbol{w_{j}}$ along the $k$-th dimension within the input text. 
	
	\noindent\textbf{3)} \textbf {An example of a Complex Vector Token Representation}
	
\begin{figure}[H]
		\vspace{-12pt}
		\centering
		\includegraphics[width=\textwidth]{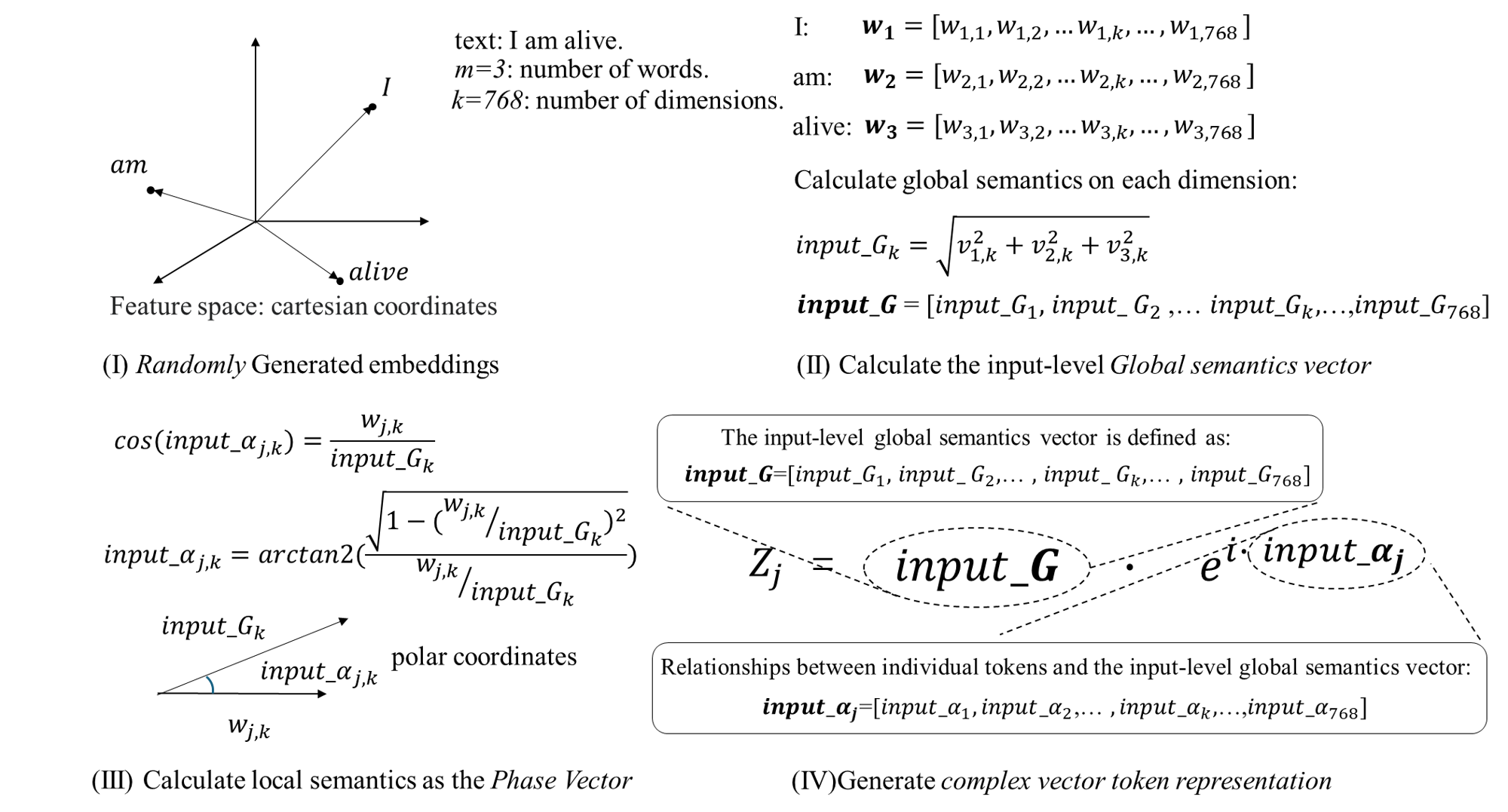}
		\caption{An example of constructing representations with \textbf{complex vector token representations}}
		\label{theory}
		\vspace{-18pt}
	\end{figure}

	Figure \ref{theory} provides an example illustrating how to convert each token in an input text ``I am alive'' into \textbf{complex vector token representations} in four steps: First, we randomly generate embeddings in a 768-dimension feature space for all three tokens in the input text; Second, we calculate the \textbf{global semantic vector} of all three token embeddings across all dimensions; Third, we compute the component ${\alpha_{j,k}}$ of the \textbf{phase vector} $\boldsymbol{\alpha_j}$ for each tokens. And finally, we combine the \textbf{global semantic vector} $\mathbf{G}$ and the \textbf{phase vector} $\boldsymbol{\alpha_j}$ to compose a \textbf{complex vector token representations} for each token $\boldsymbol{w_{j}}$ in polar coordinates.
	
	\subsection{Update Wave Representation by Superposition}\label{sec:superposition}
	\textbf{Complex vectors correspond to physical wave representations} \cite{berman2018introductory, herman2015introduction}, which allows us to use wave-based operations for more efficient \textbf{complex vector token representations} updates. We propose to develop a linear layer in our Wave network to generate two variants of the input-text level \textbf{complex vector token representation} of each token. These two variants can then be used for applying wave-based operations, such as interference and modulation to model \textbf{complex vector token representation} updates more effectively. 
	
	\subsubsection{Wave Interference} \label{waveinterference}
	From physical perspective, \textit{wave interference is a phenomenon where two coherent waves combined by adding their intensities or displacements, considering their phase difference.} In the context of generating \textbf{complex vector token representations} from input-level global semantics, as discussed in the Subsection \ref{sec:global_semantics}, we consider two variants of \textbf{complex vector token representations} for token \( \boldsymbol{w_{j}} \) as: $\mathbf{input\_Z_{j}} = \mathbf{input\_G} \cdot e^{i \cdot \boldsymbol{input\_\alpha_j}}$ and $  \mathbf{input\_Z_{j}'} = \mathbf{input\_G'} \cdot e^{i \cdot \boldsymbol{input\_\alpha_j'}}$. We use complex vectors addition to simulate wave interference \cite{openstax2024university} and obtain the combined \textbf{complex vector token representation} $\mathbf{interference\_Z_{j}}$ for token $\boldsymbol{w_{j}}$ as follows:
	\setlength{\abovedisplayskip}{3pt}  
	\setlength{\belowdisplayskip}{3pt}  
	\begin{equation}
		\begin{split}
			&\mathbf{Interference\_Z_{j}} = \mathbf{input\_Z_{j}} + \mathbf{input\_Z_{j}'}= \mathbf{input\_G} \cdot e^{i \cdot \boldsymbol{input\_\alpha_j}} + \mathbf{input\_G'} \cdot e^{i \cdot \boldsymbol{input\_\alpha_j'}}\\
			&= \left( \mathbf{input\_G} \cdot \cos(\boldsymbol{input\_\alpha_j}) + \mathbf{input\_G'} \cdot \cos(\boldsymbol{input\_\alpha_j'}) \right) \\
			&+ i \cdot \left( \mathbf{input\_G} \cdot \sin(\boldsymbol{input\_\alpha_j}) + \mathbf{input\_G'} \cdot \sin(\boldsymbol{input\_\alpha_j'}) \right)
		\end{split}
		\label{formula_interference4}
	\end{equation}
	, where the last equation is based on Euler's formula: \( e^{i\theta} = \cos(\theta) + i \sin(\theta) \). 
	
	Then, based on the definitions of the $\mathbf{input\_G}$ and $\boldsymbol{input\_\alpha_j}$ provided in the Subsection \ref{sec:global_semantics} and \ref{sec:local_semantics}, the components of Equation \ref{formula_interference4} can be expressed explicitly for the $k$-th component of the token $\boldsymbol{w_{j}}$'s \textbf{complex vector token representation} as:

	\begin{equation}
		\begin{split}
			&{Interference\_Z_{j,k}}=\sqrt{{w_{1,k}^2 + w_{2,k}^2 + \dots+w_{j,k}^2 +\dots+ w_{n,k}^2}}\cdot \frac{w_{j,k}}{\sqrt{{w_{j,k}^2 + w_{2,k}^2 + \dots+ \dots +w_{j,k}^2 +\dots+ w_{n,k}^2}}}\\
			&+\sqrt{{w_{1,k}'^2 + w_{2,k}'^2 + \dots+w_{j,k}'^2 +\dots+ w_{n,k}'^2}} \cdot \frac{w_{j,k}'}{\sqrt{{w_{j,k}'^2 + w_{2,k}'^2 + \dots+ \dots +w_{j,k}'^2 +\dots+ w_{n,k}'^2}}}\\
			&+i \cdot(\sqrt{{w_{1,k}^2 + w_{2,k}^2 + \dots+ w_{j,k}^2 +\dots+ w_{n,k}^2}})\cdot(\sqrt{1-(\frac{w_{j,k}}{\sqrt{{w_{1,k}^2 + w_{2,k}^2 + \dots +w_{j,k}^2 +\dots+ w_{n,k}^2}}})^2})\\
			&+i \cdot(\sqrt{{w_{1,k}'^2 + w_{2,k}'^2 + \dots+ w_{j,k}'^2 +\dots+ w_{n,k}'^2}})\cdot(\sqrt{1-(\frac{w_{j,k}'}{\sqrt{{w_{1,k}'^2 + w_{2,k}'^2 +  \dots +w_{j,k}'^2 +\dots+ w_{n,k}'^2}}})^2})\\
			&= w_{j,k} + w'_{j,k}+ i \cdot (\sqrt{{w_{1,k}^2 + w_{2,k}^2 + \dots+w_{j-1,k}^2 +w_{j+1,k}^2 +\dots+ w_{n,k}^2}}\\
			&+\sqrt{{w_{1,k}'^2 + w_{2,k}'^2 + \dots+w_{j-1,k}'^2 +w_{j+1,k}'^2 +\dots+ w_{n,k}'^2}})
		\end{split}
		\label{formula_interference5}
	\end{equation}

	Next, we will illustrate how the phase difference of two \textbf{complex vector token representations}, such as $ \mathbf{input\_Z_{j}}$ and $ \mathbf{input\_Z_{j}'}$, affects the overall intensity of the resulting \textbf{complex vector token representation} by analyzing their interference term. As mentioned in \cite{born2019principles, hecht2016optics}, the interference term $\text{Re}(\mathbf{input\_Z_j} \cdot \overline{\mathbf{input\_Z_j'}})$ can obtained from the square of the magnitude of ${input\_Z_{j,k}}=|\mathbf{input\_Z_j} + \mathbf{input\_Z_j'}|^2 = |\mathbf{input\_Z_j}|^2 + |\mathbf{input\_Z_j}'|^2 + 2 \cdot (\mathbf{input\_Z_j} \cdot \overline{\mathbf{input\_Z_j'}})$, which describes how the phase difference between two \textbf{complex vector token representations} determines whether they interfere constructively or destructively. $\overline{\mathbf{Input\_Z_j}}$ is the complex conjugate of $\mathbf{input\_Z_j}$, which is used to ensure that as a measurable physical quantity, the intensity is a real value \cite{griffiths2017introduction}. We can further express the interference term as:

	\begin{equation}
		\begin{split}
			2 \cdot \text{Re} (\mathbf{input\_Z_j} \cdot \overline{\mathbf{input\_Z_j'}}) &= 2 \cdot \text{Re} \left(\mathbf{input\_G} \cdot e^{i \cdot \boldsymbol{input\_\alpha_j}} \cdot \mathbf{input\_G'} \cdot e^{-i \cdot \boldsymbol{input\_\alpha_j'}}\right) \\
			&= 2 \cdot \mathbf{input\_G} \cdot \mathbf{input\_G'} \cdot \cos(\boldsymbol{input\_\alpha_j} - \boldsymbol{input\_\alpha_j'})
		\end{split}
		\label{interferenceitem2}
	\end{equation}

	Equation \ref{interferenceitem2} demonstrates that the cosine value of the phase difference directly determines the interference result.
	
	\subsubsection{Wave Modulation}\label{sebsection:wavemodulation}
	From a physical perspective, \textit{wave modulation is the process of varying one or more properties of a periodic waveform, called the carrier signal, with a separate signal, which contains the information to be transmitted.} From a signal processing perspective, \textit{adjusting the amplitude of a carrier wave in accordance with the input signal is referred to as amplitude modulation \cite{georgi1993physics}.} Similarly, \textit{adjusting the phase of a carrier wave in accordance with the input signal is referred to as phase modulation \cite{phasemod}.} Both amplitude modulation and phase modulation can be achieved through the multiplication of complex vectors representing waves \cite{CRECRAFT2002200, PURSLEY200223, james2003digital}. 
	
	In the context of generating \textbf{complex vector token representations} from input-level global semantics, as discussed in the Subsection \ref{sec:global_semantics}, we consider two variants of \textbf{complex vector token representations} for token \( \boldsymbol{w_{j}} \) as: $\mathbf{input\_Z_{j}} = \mathbf{input\_G} \cdot e^{i \cdot \boldsymbol{input\_\alpha_j}}$ and $  \mathbf{input\_Z'_{j}} = \mathbf{input\_G'} \cdot e^{i \cdot \boldsymbol{input\_\alpha_j'}}$. We use complex vectors multiplication to simulate wave modulation \cite{openstax2024university} and obtain the combined \textbf{complex vector token representation} $\mathbf{modulation\_Z_{j}}$ for token $\boldsymbol{w_{j}}$ as follows:

	\begin{equation}
		\begin{split}
			&\mathbf{Modulation\_Z_{j}} = \mathbf{input\_Z_{j}} \cdot \mathbf{input\_Z_{j}'}= \mathbf{input\_G} \cdot e^{i \cdot \boldsymbol{input\_\alpha_j}} \cdot  \mathbf{input\_G'} \cdot e^{i \cdot \boldsymbol{input\_\alpha_j'}}  \\
			&= \mathbf{input\_G} \cdot \mathbf{input\_G'} \cdot e^{i \cdot \boldsymbol{input\_\alpha_j} + \boldsymbol{input\_\alpha_j'}}\\
			&=\mathbf{input\_G} \cdot \mathbf{input\_G'} \cdot \cos(\boldsymbol{input\_\alpha_j} + \boldsymbol{input\_\alpha_j}')\\
			&+ i \cdot \mathbf{input\_G} \cdot \mathbf{input\_G'} \cdot \sin(\boldsymbol{input\_\alpha_j}+ \boldsymbol{input\_\alpha_j'})\\
		\end{split}
		\label{formula_modulation1}
	\end{equation}

	Here, the amplitude modulation term is represented by $\mathbf{input\_G} \cdot \mathbf{input\_G'}$. And the phase modulation term is expressed as $\boldsymbol{input\_\alpha_j} + \boldsymbol{input\_\alpha_j'}$. Then, based on the definitions of the $\mathbf{input\_G}$ and $\boldsymbol{input\_\alpha_j}$ provided in the Subsection \ref{sec:global_semantics} and \ref{sec:local_semantics}, the components of Equation \ref{formula_modulation1} can be expressed explicitly for the $k$-th component of the token $\boldsymbol{w_{j}}$'s \textbf{complex vector token representation} as:

\begin{equation}
	\begin{split}
		&{Modulation\_Z_{j,k}}=\sqrt{{w_{1,k}^2 + w_{2,k}^2 + \dots+w_{j,k}^2 +\dots+ w_{n,k}^2}} \cdot \sqrt{{w_{1,k}'^2 + w_{2,k}'^2 + \dots+w_{j,k}'^2 +\dots+ w_{n,k}'^2}}\\
		& \cdot \left({\frac{w_{j,k}}{\sqrt{w_{1,k}^2 + w_{2,k}^2 + \dots +w_{j,k}^2 + \dots +w_{n,k}^2}} \cdot \frac{w_{j,k}'}{\sqrt{w_{1,k}'^2 + w_{2,k}'^2 + \dots +w_{j,k}'^2 + \dots +w_{n,k}'^2}}}\right.\\
		&\left.-\sqrt{1-(\frac{w_{j,k}}{\sqrt{{w_{1,k}^2 + w_{2,k}^2 + \dots +w_{j,k}^2 + \dots +w_{n,k}^2}}})^2} \cdot \sqrt{1-(\frac{w_{j,k}'}{\sqrt{{w_{1,k}'^2 + w_{2,k}'^2 + \dots +w_{j,k}'^2 +\dots+ w_{n,k}'^2}}})^2} \right)\\
		&+i \cdot \sqrt{{w_{1,k}^2 + w_{2,k}^2 + \dots +w_{j,k}^2 + \dots +w_{n,k}^2}} \cdot \sqrt{{w_{1,k}'^2 + w_{2,k}'^2 + \dots+w_{j,k}'^2 +\dots+ w_{n,k}'^2}}\\
		& \cdot \left(\sqrt{1-(\frac{w_{j,k}}{\sqrt{{w_{1,k}^2 + w_{2,k}^2 + \dots +w_{j,k}^2 + \dots +w_{n,k}^2}}})^2} \cdot  \frac{w_{j,k}'}{\sqrt{{w_{1,k}'^2 + w_{2,k}'^2 + \dots+ w_{j,k}'^2 +\dots+ w_{n,k}'^2}}}\right.\\
		&\left.+\frac{w_{j,k}}{\sqrt{w_{1,k}^2 + w_{2,k}^2 + \dots +w_{j,k}^2 + \dots +w_{n,k}^2}} \cdot \sqrt{1-(\frac{w_{j,k}'}{\sqrt{{w_{1,k}'^2 + w_{2,k}'^2 + \dots +w_{j,k}'^2 +\dots+ w_{n,k}'^2}}})^2}\right) \\
		&= (w_{j,k}\cdot w_{j,k}'-\sqrt{{w_{1,k}^2 + w_{2,k}^2 + \dots+w_{j-1,k}^2  +w_{j+1,k}^2 + \dots +w_{n,k}^2}} \\
		&\cdot \sqrt{{w_{1,k}'^2 + w_{2,k}'^2 + \dots+w_{j-1,k}'^2 +w_{j+1,k}'^2 +\dots+ w_{n,k}'^2}})\\
		&+ i \cdot  \left(w_{j,k}' \cdot \sqrt{{w_{1,k}^2 + w_{2,k}^2 + \dots+w_{j-1,k}^2 + w_{j+1,k}^2 + \dots +w_{n,k}^2}}\right.\\
		&\left.+ (w_{j,k} \cdot \sqrt{{w_{1,k}'^2 + w_{2,k}'^2 + \dots+w_{j-1,k}'^2 +w_{j+1,k}'^2 +\dots + w_{n,k}'^2}}\right)
	\end{split}
	\label{formula_modulation555}
\end{equation}

	\section{Single-layer and Multi-layer Architecture of the Wave Network}

\noindent\textbf{a)} \textit{Single layer architecture:} 

\begin{figure}[ht]
	\vspace{-15pt}
	\centering
	\begin{subfigure}[b]{0.5\columnwidth} 
		\centering
		\includegraphics[width=\textwidth]{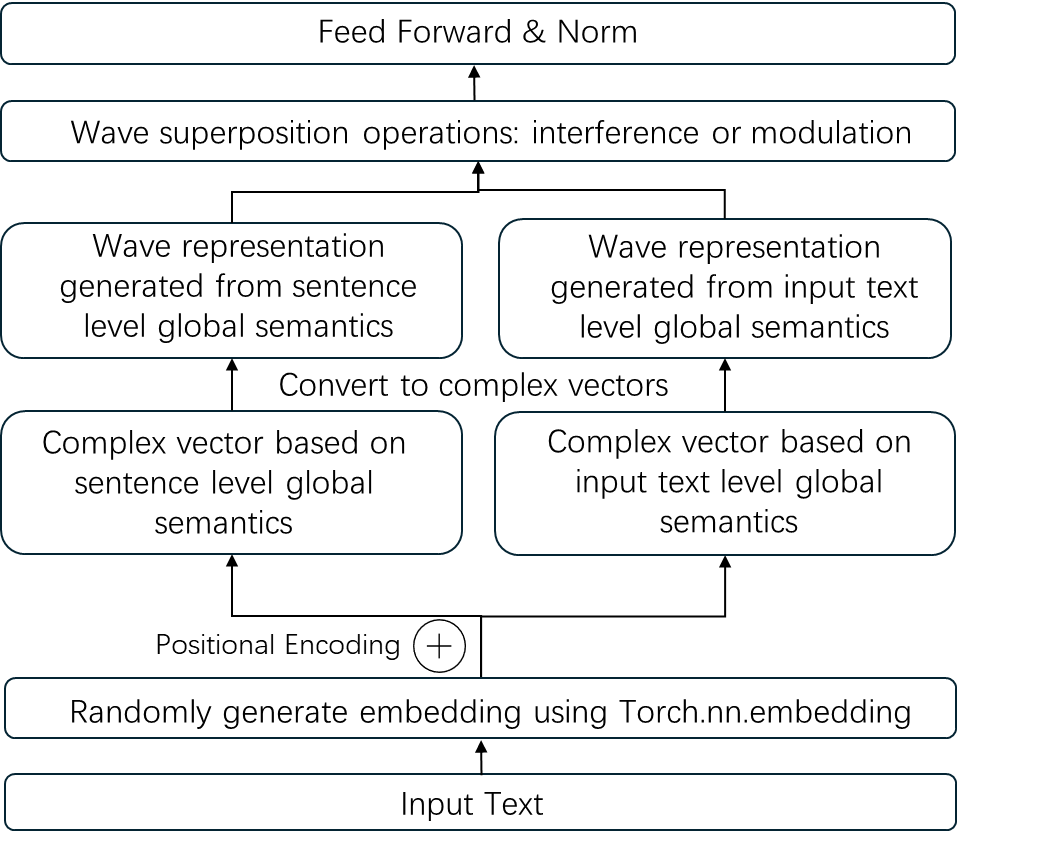}
		\caption{Single layer of Wave network}
		\label{fig:mtov}
		\vspace{-5pt}
	\end{subfigure}
	\hspace{0.05\columnwidth} 
	\begin{subfigure}[b]{0.25\columnwidth} 
		\centering
		\includegraphics[width=\textwidth]{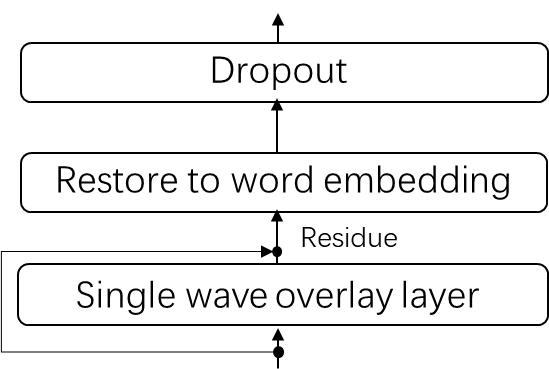}
		\caption{One-layer block of Wave network}
		\label{fig:phitov}
		\vspace{-5pt}
	\end{subfigure}
	\caption{Single layer and multi-layer design of the Wave Network}
	\label{fig:comparison}
	\vspace{-20pt}
\end{figure}

To implement the operations of generating and updating \textbf{complex vector token representations}, we designed a deep learning model called the Wave network. Figure \ref{fig:mtov} illustrates the layer-level design of the Wave network. Starting with an input text, we generate initial embeddings for each token by randomly assigning values across 768 dimensions. These initial embeddings serve as the basis for generating a single \textbf{wave representation} representing input-level global semantics, with positional encoding included. Using a linear layer, we create two variants of this \textbf{wave representation}, which are then modulated through multiplication. The resulting representation is passed through a feed-forward layer and a normalization layer. Finally, the processed representations are used as the foundation for subsequent classification tasks.\\
\small
\normalsize
\noindent\textbf{b)} \textit{Multi-layer architecture:} 

Figure \ref{fig:phitov} shows the structure of one block in a multi-layer Wave network, which mathematical expressions can be represented as:
\small
\vspace{-5pt}
\begin{equation}
	\mathbf{W_{n+1}} = \mathbf{W_n} + D\left(g\left(f_{n+1}(\mathbf{W_n})\right)\right)
\end{equation}
\normalsize
Where $\mathbf{W_n}$ is the output of the $n$-th layer, $f_{n+1}(\mathbf{W_n})$ denotes the wave overlay of the $(n+1)$-th layer, $g(x)$ convert complex number to token embedding, and $D(x)$ the dropout operation applied to the input $x$.
	
	\section{Experiments and Results}
\begin{figure*}[h]
		\centering
		\includegraphics[width=1\linewidth]{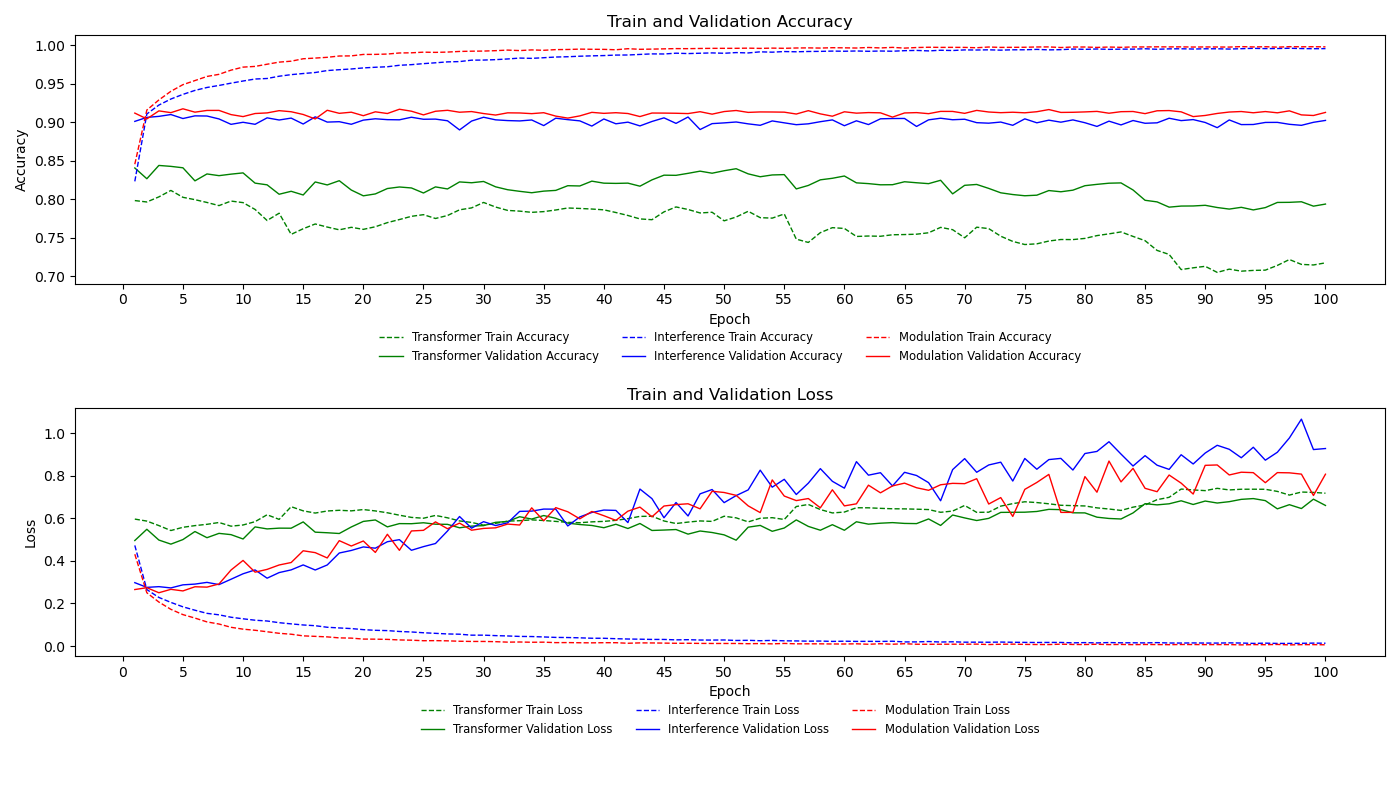}
		\caption{Comparison between Wave Network and Transformer}
		\label{fig:additional_figure}
	\end{figure*}

	The parameters across all experiments are kept as consistent as possible. The learning rate for both the Wave network and Transformer learning rate is set to 1e-3, while for BERT, it is set to 2e-5\cite{sun2019fine}. The batch size varies depending on the task. In the resource utilization comparison experiments, all models use a batch size of 64. In the accuracy comparison experiments, the batch size is 64 for the Wave network and Transformer, and 32 for BERT\cite{sun2019fine}. In the gradient comparison and embedding independence experiments, all three models use a batch size of 32. All models are trained or fine-tuned for four epochs. In the fast convergence experiments, we use 500 batches and record the accuracy on the test set every ten batches. The Wave network and Transformer use a single-layer design; the Wave network generates initial token embeddings randomly using torch.nn.embedding in Pytorch\cite{paszke2019pytorch}, while the Transformer uses token embeddings from pre-trained BERT. For BERT experiments, the base pre-trained model is fine-tuned. In Table \ref{fig:additional_figure} and \ref{tab:performance_comparison_2}, VRAM usage is measured in gigabytes(GB), and the time consumed per epoch(for both training and fine-tuning) is measured in seconds. All preliminary experiments are conducted on AG News\cite{bib16}, DBpedia14\cite{ref17auer2007dbpedia}, and IMDB\cite{ref18maas2011learning} datasets, with an 80/20 training-validation split. Each data has a separated test subset. 
	
	\begin{table*}[h]
		\centering
		\begin{tabular}{>{\centering\arraybackslash}m{6cm}
				>{\centering\arraybackslash}m{1.5cm}
				>{\centering\arraybackslash}m{2cm}
				>{\centering\arraybackslash}m{1.8cm}
				>{\centering\arraybackslash}m{1.3cm}}
			\toprule
			Model & Data sets & Accuracy & VRAM(GB) & Time(s)  \\
			\midrule
			Wave network (Interference) & AG News & 90.36\% & 0.30 & 146.90 \\
			Wave network (Modulation) & AG News & 91.29\% & 0.32 & 147.02 \\
			Transformer & AG News & 71.68\% & 0.85 & 173.25 \\
			\bottomrule
		\end{tabular}
		\caption{Performance Comparison between Wave Network and Transformer on AG News}
		\label{tab:performance_comparison}
	\end{table*}
	\begin{table*}[h]
		\centering
		\begin{tabular}{>{\centering\arraybackslash}m{6cm}
				>{\centering\arraybackslash}m{1.5cm}
				>{\centering\arraybackslash}m{2cm}
				>{\centering\arraybackslash}m{1.8cm}
				>{\centering\arraybackslash}m{1.3cm}}
			\toprule
			Model & Data sets & Accuracy & VRAM(GB) & Time(s) \\
			\midrule
			Wave network (Interference) & AG News & 90.91\% & 0.30 & 146.90 \\
			Wave network (Modulation) & AG News & 91.66\% &  0.32 & 147.02 \\
			BERT base & AG News & 94.64\% & 1.28 & 1034.99 \\
			Wave network (Interference) & DBpedia14 & 97.93\% & 0.30 & 979.78 \\
			Wave network (Modulation) & DBpedia14 & 98.05\% & 0.30 & 991.15 \\
			BERT base & DBpedia14 & 99.28\% & 1.27 & 2734.76 \\
			Wave network (Interference) & IMDB & 87.00\% & 0.37 & 119.57 \\
			Wave network (Modulation) & IMDB & 87.02\% & 0.37 & 119.96 \\
			BERT base & IMDB & 93.94\% & 1.27 & 220.46 \\
			\bottomrule
		\end{tabular}
		\caption{Performance Comparison between Wave Network and BERT on Various Data Sets}
		\label{tab:performance_comparison_2}
	\end{table*}
	
	As shown in Table \ref{tab:performance_comparison} and Table \ref{tab:performance_comparison_2} for the AG News dataset, compared to the single-layer Transformer, a single-layer Wave network achieved 90.36\% accuracy with wave interference and 91.29\% with wave modulation, outperforming a single Transformer layer by 18.68\% and 19.61\%, respectively, and approaching the BERT base model's 94.64\%.  At the same time, the Wave network reduced video memory usage and training time by 77.34\% and 85.62\% during wave modulation compared to BERT base.
	
	\section{Discussion}\label{sec3}
	The Wave Network introduces an efficient token representation method that uses complex vectors to encode global and local text semantics, allowing it to perform competitively on NLP tasks with significantly reduced computational demands. By simulating wave interference and modulation, the model achieves high accuracy with much lower VRAM usage and training time compared to larger models like BERT base. This efficiency not only highlights its potential for resource-limited devices but also paves the way for scalable NLP applications that can be deployed in real-time environments, from mobile devices to embedded systems without sacrificing performance.
	\bibliography{reference}

\end{document}